\documentclass[sigconf,nonacm]{acmart}

\AtBeginDocument{%
  }

\setcopyright{none}

\usepackage{multirow}
\title{TriCLIP-3D: A Unified Parameter-Efficient Framework for Tri-Modal 3D Visual Grounding based on CLIP}

\author{
    Fan Li\textsuperscript{1,2†}, 
    Zanyi Wang\textsuperscript{1,2†}, 
    Zeyi Huang\textsuperscript{4}, 
    Guang Dai\textsuperscript{2}, 
    Jingdong Wang\textsuperscript,
    Mengmeng Wang\textsuperscript{3,2,*} \\[1ex]
    \textsuperscript{1}Xi'an Jiaotong University,
    \textsuperscript{2}SGIT AI Lab,
    \textsuperscript{3}Zhejiang University of Technology,
    \textsuperscript{4}Huawei
}
\thanks{* Corresponding author. \\† These authors contributed equally to this work completed during internships at SGIT AI Lab, State Grid Corporation of China.}

\begin{document}

\begin{abstract}
3D visual grounding allows an embodied agent to understand visual information in real-world 3D environments based on human instructions, which is crucial for embodied intelligence. Existing 3D visual grounding methods typically rely on separate encoders for different modalities (e.g., RGB images, text, and 3D point clouds), resulting in large and complex models that are inefficient to train. While some approaches use pre-trained 2D multi-modal models like CLIP for 3D tasks, they still struggle with aligning point cloud data to 2D encoders. As a result, these methods continue to depend on 3D encoders for feature extraction, further increasing model complexity and training inefficiency. In this paper, we propose a unified 2D pre-trained multi-modal network to process all three modalities (RGB images, text, and point clouds), significantly simplifying the architecture. By leveraging a 2D CLIP bi-modal model with adapter-based fine-tuning, this framework effectively adapts to the tri-modal setting, improving both adaptability and performance across modalities. Our Geometric-Aware 2D-3D Feature Recovery and Fusion (GARF) module is designed to fuse geometric multi-scale features from point clouds and images. We then integrate textual features for final modality fusion and introduce a multi-modal decoder to facilitate deep cross-modal understanding. Together, our method achieves unified feature extraction and fusion across the three modalities, enabling an end-to-end 3D visual grounding model. Compared to the baseline, our method reduces the number of trainable parameters by approximately 58\%, while achieving a 6.52\% improvement in the 3D detection task and a 6.25\% improvement in the 3D visual grounding task.

% 3D visual grounding allows an embodied agent to understand visual information in real-world 3D environments based on human instructions, which is crucial for embodied intelligence. Current methods often use separate encoders for different modalities (e.g., RGB images, text, and 3D point clouds), leading to large and complex models that are inefficient to train. Although some approaches employ pre-trained 2D multi-modal models like CLIP for 3D tasks, they face challenges in aligning point cloud data with 2D encoders, necessitating the use of 3D encoders and further complicating the model. In this paper, we propose a unified 2D pre-trained multi-modal network to process all three modalities—image, text, and point cloud, thus simplifying the architecture. By utilizing a 2D CLIP bi-modal model with adapter-based fine-tuning, our framework effectively adapts to a tri-modal setting, enhancing adaptability and performance across modalities. Our Geometric-Aware 2D-3D Feature Recovery and Fusion (GARF) module fuses geometric multi-scale features from point clouds and images, and we integrate textual features for final modality fusion, introducing a multi-modal decoder to facilitate deep cross-modal understanding. This approach achieves unified feature extraction and fusion across the three modalities, enabling an end-to-end 3D visual grounding model. Compared to the baseline, our method reduces trainable parameters by approximately 58\% and achieves a 6.52\% improvement in the 3D detection task and a 6.25\% improvement in the 3D visual grounding task.
\end{abstract}

%    <concept>
%        <concept_id>10010147.10010178.10010224</concept_id>
%        <concept_desc>Computing methodologies~Computer vision</concept_desc>
%        <concept_significance>500</concept_significance>
%        </concept>

%    <concept>
%        <concept_id>10010147.10010178</concept_id>
%        <concept_desc>Computing methodologies~Artificial intelligence</concept_desc>
%        <concept_significance>500</concept_significance>
%        </concept>
%  </ccs2012>

% \ccsdesc[500]{Computing methodologies~Computer vision}

% \ccsdesc[500]{Computing methodologies~Artificial intelligence}



% \received{11 April 2025}
% \received[revised]{TBD}  % 待定，稍后更新
% \received[accepted]{TBD}  % 待定，稍后更新


\maketitle

\section{Introduction}

Recently, egocentric 3D perception tasks have emerged as a critical research area in embodied intelligence. Among 3D tasks, multi-modal 3D visual grounding has gained widespread attention. This is because it needs an embodied agent not only to have strong localization and comprehension abilities in 3D scenes but also to accurately understand human language descriptions. This requires strong multi-modal understanding from the embodied agent.

\begin{figure}[!t]
    \centering
    \includegraphics[width=\linewidth]{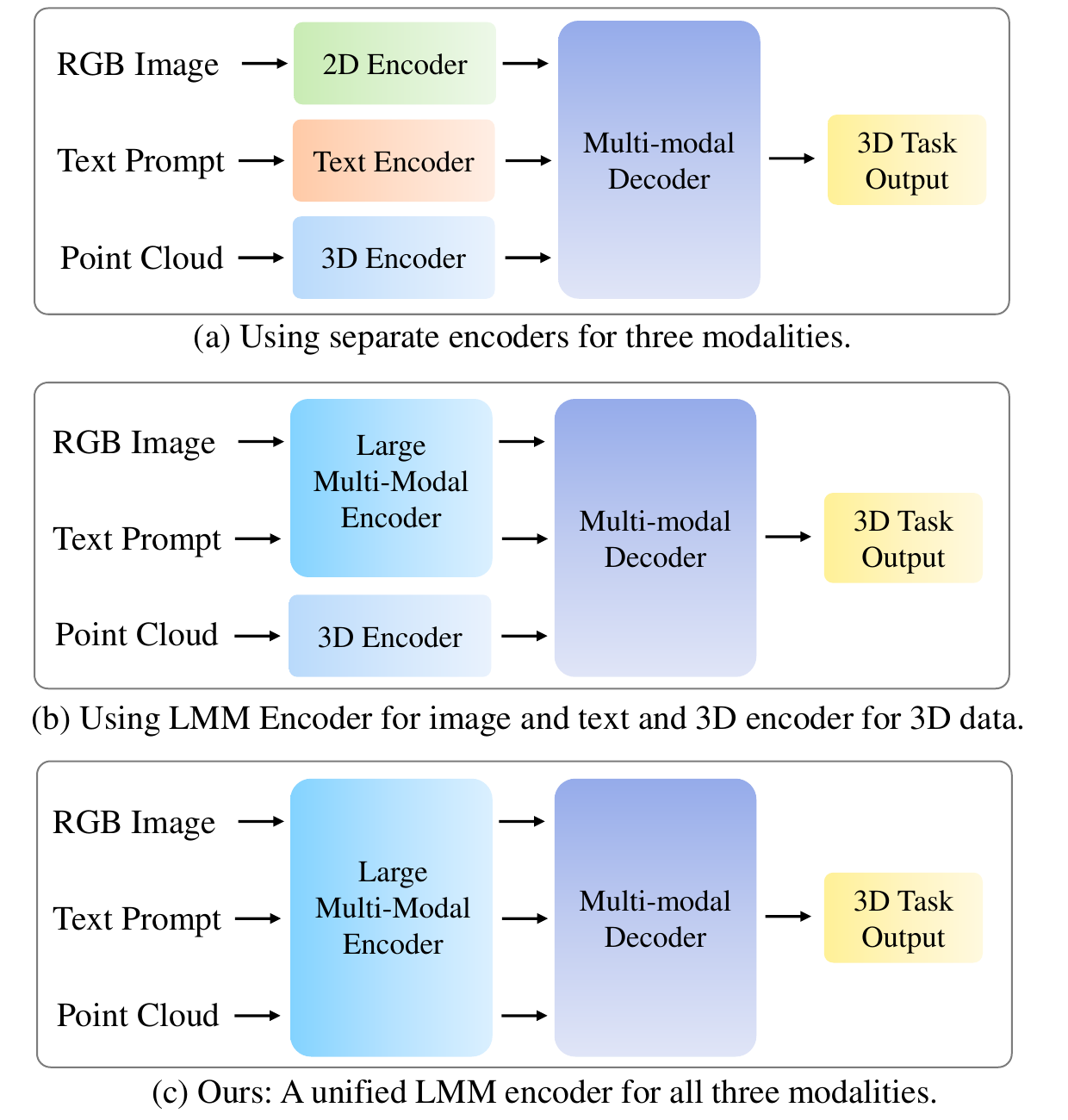}
    \caption{Comparison of different encoder architectures: (a) Separate encoders for each modality. (b) Large multi-modal encoder for image and text, with a separate 3D encoder. (c) Our approach: a unified large multi-modal encoder for all three modalities.}
    \label{introduction}
\end{figure}

Current advancements in 3D visual grounding typically follow two main paradigms for multi-modal processing. First, as shown in Figure \ref{introduction}(a), existing methods\cite{embodiedscan,multi-trans,wildrefer,jointdown} typically use separate encoders to extract features from various modalities, such as text, images and point clouds. These extracted features are then fused and fed into a multi-modal decoder to output the 3D location of the corresponding object. This fragmented design has two critical limitations: First, isolated feature extraction creates inherent misalignment between 2D image pixels, 3D points, and linguistic tokens, forcing the decoder to handle incompatible feature spaces. Second, these encoders (e.g. CNNs for images, transformers for text, 3D networks for point clouds) lead to parametric redundancy and training inefficiency. Although some approaches\cite{ulip2, clip2point, clipgoes3d, ghoseclip, pointclip} use pre-trained 2D multi-modal models (e.g. CLIP\cite{clip}) for 3D tasks, as shown in Figure \ref{introduction}(b), they still rely on 3D backbone networks (such as PointNet++\cite{pointnet++}, PointGroup\cite{pointgroup}) to extract point-cloud features. However, 3D point cloud models often require significantly more parameters than standard 2D models, leading to higher computational costs. Moreover, the sparse and irregular structure of point clouds makes them incompatible with pre-trained image-text models, which are designed to process dense and pixel 2D data. These fundamental differences in data structure and model design force researchers to use specialized 3D backbone to extract point-cloud features.

To address the mentioned issues, as shown in Figure \ref{introduction}(c), we propose TriCLIP-3D, a unified multi-modal fusion framework that processes multi-view RGB images, point clouds, and text prompts through a pre-trained CLIP for 3D tasks. Our primary aim is to incorporate a 2D pre-trained multi-modal CLIP as a unified encoder for 3D tasks through model fine-tuning, accommodating inputs from point cloud, image and text modalities. Notably, both point clouds and images share a single CLIP Vision Transformer (ViT) model for feature extraction. In this way, we don't need an additional 3D network for point-cloud feature extraction, resulting in a simplified tri-modal feature extraction framework. However, due to the sparsity of point cloud samples and the inherent domain differences from images, directly inputting point clouds into a 2D CLIP model and fine-tuning it presents significant challenges. Inspired by EPCL\cite{epcl}, we patch up point clouds to create patches. These patches are embedded and fed into the CLIP image encoder to extract features. The key reason why point clouds can effectively adapt to the CLIP image encoder lies in the structural similarity between point cloud representations and image patches at the sequential encoding level. And the knowledge learned from CLIP helps point clouds focus on the similar semantic regions as images do. We also streamline adapter design by adopting a unified strategy. Each modality’s inputs are directed to specific adapters for targeted fine-tuning during model propagation. This ensures optimal adjustment for each modality, enhancing overall model performance and extending the dual-modality feature extraction network to a tri-modality setup.

In addition, we observed that directly fusing CLIP-extracted point cloud and image tokens results in degraded cross-modal geometric consistency. This issue arises from the lack of explicit spatial constraints, such as perspective projection consistency in the CLIP encoding process, leading to misaligned geometric-semantic correlations and poor fusion performance. To address this, we propose the Geometric-Aware 2D-3D Feature Recovery and Fusion (GARF) module. Initially, we recover the features extracted from CLIP for both point clouds and image sequences into 3D sparse tensors and 2D feature maps, generating multi-scale features. By projecting the point cloud onto the image features, we use the Adaptive Point-Image Fusion Module (APIF) for dynamic fusion. This method not only filters out irrelevant features but also enhances feature complementarity by effectively combining spatial and contextual information from both modalities.

Together, we validated our approach on the EmbodiedScan benchmark, focusing on the tasks of 3D detection and 3D visual grounding. Compared to EmbodiedScan\cite{embodiedscan}, our model reduces trainable parameters by 58\%, while achieving a 6.52\% improvement in accuracy for 3D detection and a 6.25\% improvement for 3D visual grounding.

Our main contributions can be summarized as follows.
\begin{itemize}
    \item 
    We propose TriCLIP-3D, a unified tri-modal feature extraction framework leveraging a single pre-trained CLIP model to encode text, multi-view images, and point clouds. This approach uniquely utilizes the same CLIP visual encoder for both images and point clouds, eliminating the need for separate 3D network backbones.
    \item We propose the Geometric-Aware 2D-3D Feature Recovery and Fusion (GARF) module, which enhances cross-modal geometric consistency by recovering spatial context and adaptively fusing features based on 3D-to-2D projection. This approach leads to significantly improved feature fusion performance for 3D tasks.
    \item Compared to baseline, our method significantly reduces trainable parameters by 58\% and achieves notable accuracy improvements of 6.52\% in 3D detection and 6.25\% in 3D visual grounding on the EmbodiedScan benchmark.
\end{itemize}

\begin{figure*}[!t]
    \centering
    \includegraphics[width=\linewidth]{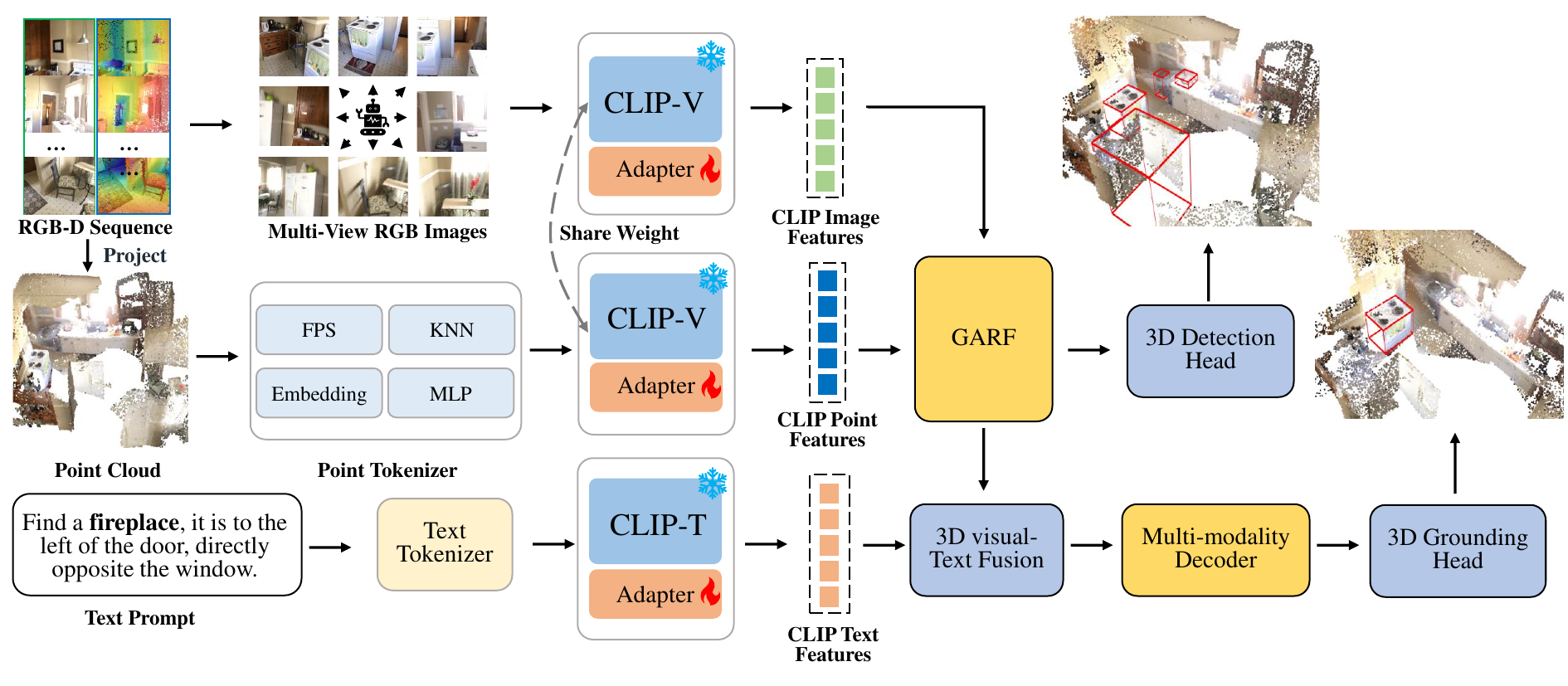}
    \caption{Overview of our framework. The proposed network architecture integrates multiple modalities, including RGB images, depth data, and textual information, to perform 3D object detection or 3D visual grounding task. It adopts a unified CLIP-Visual encoder with shared parameters for both the point cloud and image branches. During training, the CLIP-Visual encoder and CLIP-Text encoder remain frozen, while trainable Adapters are introduced to fine-tune CLIP for 3D tasks. }
    \label{fig:method1}
\end{figure*}

\section{Related Work}
\subsection{3D visual grounding}
3D visual grounding is the task of precisely identifying and localizing objects described in language instructions within 3D scenes. Early studies like ReferIt3D\cite{referit3d} and ScanRefer\cite{scanrefer} established 3D visual grounding benchmarks using ScanNet\cite{scannet} data. The majority of 3D visual grounding methods\cite{free,3djcg,multi-trans,mikasa,unit3d,dreftransformer,languagerefer} use two-stage model, they first train a 3D detector to find proposal regions, then combine these regions with text features through interaction to get the final 3D visual grounding results. Some one-stage methods\cite{3dvg,eda,dense,lamm,wu3d,bottom} usually
employ multi-modal architectures to directly output the results of 3D visual grounding. To enhance 3D spatial awareness, SAT\cite{sat} incorporates 2D semantics as additional input during training. This approach aids 3D visual grounding by leveraging the auxiliary objectives of 2D visual grounding. And also, LAR \cite{LAR} uses a 2D Synthetic Images Generator (SIG) to create multi-view 2D images from 3D point clouds, which are then integrated into a multi-modal transformer-based architecture to improve the grounding performance. FFL-3DOG\cite{free} leverages language and visual scene graphs to facilitate the alignment of features between linguistic inputs and point cloud data. \cite{open3d} utilizes large language models to overcome the limitations of traditional methods that require extensive annotations for zero-shot open-vocabulary 3D visual grounding. However, these methods either support only point cloud and text prompt inputs or multi-view image inputs. In contrast, our architecture supports tri-modal inputs, including multi-view images, point clouds, and text prompts.

\subsection{Vision foundation model for 3D task}
Due to the powerful generalization capabilities of large multi-modal models such as CLIP\cite{clip} and Slip\cite{slip}, there is a growing trend in research to transfer these pretrained 2D multi-modal models to 3D tasks. CLIP2Point\cite{clip2point} explores the adaptation of the CLIP model for point cloud classification by leveraging pre-training on image-depth data. ULIP\cite{ulip} stands out as an early effort in developing triplets that integrate 3D point clouds, images, and language for 3D Classification. However, ULIP still utilizes a 3D encoder to extract point-cloud features, which are then aligned with image and text features extracted by CLIP. UNI3D\cite{unit3d} employs a unified transformer-based architecture designed to integrate and align 3D point-cloud features with image-text features. However, UNI3D still utilizes a learnable Vision Transformer (ViT) for extracting 3D features. Cross3DVG\cite{cross3dvg} uses CLIP model to extract multi-view image features to boost grounding effects. But it still uses VoteNet\cite{votenet:} to take a point cloud as input and to predict object proposals within the scene. Unlike these methods, our TriCLIP-3D approach does not rely on an additional 3D network to extract point-cloud features. Instead, it utilizes a unified CLIP model to extract features from multi-view images, point clouds, and text prompts.

\section{Methods}
In this section, we introduce our 3D visual grounding framework, which primarily consists of TriCLIP-3D encoder and several multi-modal fusion modules. Each subsection will detail the individual methodologies.
% {\bfseries Your document will be returned to you for revision if
%   modifications are discovered.}
\subsection{Overall Framework}
As shown in Figure \ref{fig:method1}, our framework consists of three main branches. We innovatively leverage the pretrained 2D multi-modal CLIP model to extract features from three modalities, such as point cloud, Multi-View RGB images and text. First, the RGB-D sequence aggregates multi-view RGB images, which are extracted by the visual encoder of CLIP to obtain multi-view image features. Second, the RGB-D sequence uses camera parameters to project the depth map into 3D point cloud scenes, after processing by the point tokenizer, 3D point clouds are fed into the unified CLIP visual encoder to extract point-cloud features. Third, the text prompt is processed by the CLIP text tokenizer and is extracted by the CLIP text encoder to obtain text features. After feature extraction from the three modalities is completed, the image features and point-cloud features are fused to obtain point cloud-image fused feature through GARF. In the 3D detection task, this fusion feature directly outputs multiple 3D bounding boxes (3D BBOX) from a 3D Detection Head. In the 3D visual grounding task, the point cloud-image fused feature is further combined with the text features to obtain a three-modal fused feature, which is then input into a multi-modal decoder. Finally, the decoder outputs the 3D bounding box of the main object described in the text prompt.

\subsection{TriCLIP-3D Encoder}
We use a unified pretrained CLIP model to extract features from three modalities: multi-view images, 3D point clouds, and text. Both multi-view images and 3D point clouds share the same CLIP visual encoder, while the text is processed using the CLIP text encoder. During training, the CLIP model is frozen, significantly improving training efficiency. To enable better transfer of the CLIP 2D vision-language model to 3D tasks, we further fine-tune the CLIP model using residual adapter.

\textbf{(1) Multi-View Images Feature Extraction.} The extraction of multi-view image features is a crucial step in our framework, as it allows for a comprehensive understanding of the 3D scene from different perspectives. Initially, the multi-view images undergo data preprocessing to form a tensor $I\in R^{B\times Nums\times C\times H\times W} $, $Nums$ represents the number of image views. Given that the original CLIP model is designed for single-image inputs, it does not natively support the direct processing of multiple images simultaneously. To address this limitation, we aggregate the multi-view images into a single batch $I_{N}  \in R^{B * Nums\times C\times H\times W}$. This aggregation enables the processing of multiple views as a unified input, which is then fed into the CLIP Vision Transformer (ViT). Then the ViT processes these aggregated inputs to extract CLIP multi-view image features $f_{mv} \in R^{(B* Nums)\times L\times D_{1}}$.

\textbf{(2) Point-Cloud Feature Extraction.} In recent years, numerous point-cloud feature extraction networks based on the transformer architecture have been proposed. These networks typically begin by tokenizing the point cloud. Given a 3D Point Cloud Scene $P \in R^{S\times 3}$, following EPCL\cite{epcl}, we first use 3D Minkowski Convolution\cite{meconv} to extract point-cloud features $S_{1} =SparseTensor(F_{1} ,U_{1} )$, $F_{1} \in R^{N_{1}\times D'}$ represents point feature, $U_{1} \in R^{N_{1}\times 3} $ represents the three-dimensional coordinates, then apply the Farthest Point Sampling (FPS) algorithm to the point cloud. The FPS algorithm selects the most distant points in the set, ensuring that the sampled points are well-distributed in the 3D space. Next, we group $K$ points around each center using the K-Nearest Neighbourhood (KNN) algorithm. This process generates $M$ patches, where each patch represents a local region of the point cloud, capturing spatial relationships between the neighboring points. After grouping, we embed the 3D coordinates of the point cloud, and the point cloud patches are passed through MLP to get tokens $P_{T}  \in R^{M\times D} $, then they are fed into the CLIP visual encoder to get final CLIP point feature $f_{p}  \in R^{B\times L \times D_{1}} $. These processes can be summarized by the following equations:
$$
f_{p}  = MLP(Emb(Knn(Fps(P)))) \eqno{(1)}
$$

\textbf{(3) Text Feature Extraction.} For the 3D visual grounding task, we tokenize the text prompt and extract features using the CLIP text encoder. Unlike traditional language transfer tasks with CLIP, the text prompt in 3D visual grounding may contain multiple objects. Therefore, using the original global feature output of CLIP can lead to ambiguity regarding the referenced objects. In our framework, we use the feature sequence for each text token $f_{t}  \in R^{B\times L \times D_{2} }$, to enable proper alignment with the visual features in the feature fusion steps.

\textbf{(4) Residual Adapter.} To better transfer the pre-trained 2D CLIP model to 3D tasks, we introduce a residual adapter layer into the transformer architecture of the CLIP model. This allows us to fine-tune the CLIP model, improving its performance on 3D tasks. Specifically, we inserted residual adapters into the odd-numbered layers of both the ViT component and the text transformer of CLIP. Let $x \in R^{B\times L \times D }$ be the output sequence of a transformer block. The adapter layer is a lightweight module consisting of two fully connected layers, specifically defined as follows:
$$
x_{1} = GeLU(W_{1}*x+ b_{1} )  \eqno{(2)}
$$
$$
x_{out} = x + (W_{2}*x_{1}  +b_{2})  \eqno{(3)}
$$
where $W_{1}, W_{2}$ represents the weights and $ b_{1}, b_{2}$ denotes the bias.

Specifically, we use a zero-weight constant initialization strategy to enhance the stability of fine-tuning and prevent disruption of the original model. During the forward pass of the model, the introduction of the residual adapter layer enables the model to selectively adjust its representation based on the input data modality (RGB images, point cloud or text data), thereby facilitating the extraction of features for all three modalities.

\subsection{Geometric-Aware 2D-3D Feature Recovery and Fusion} 
\begin{figure}[!t]
    \centering
    \includegraphics[width=\linewidth]{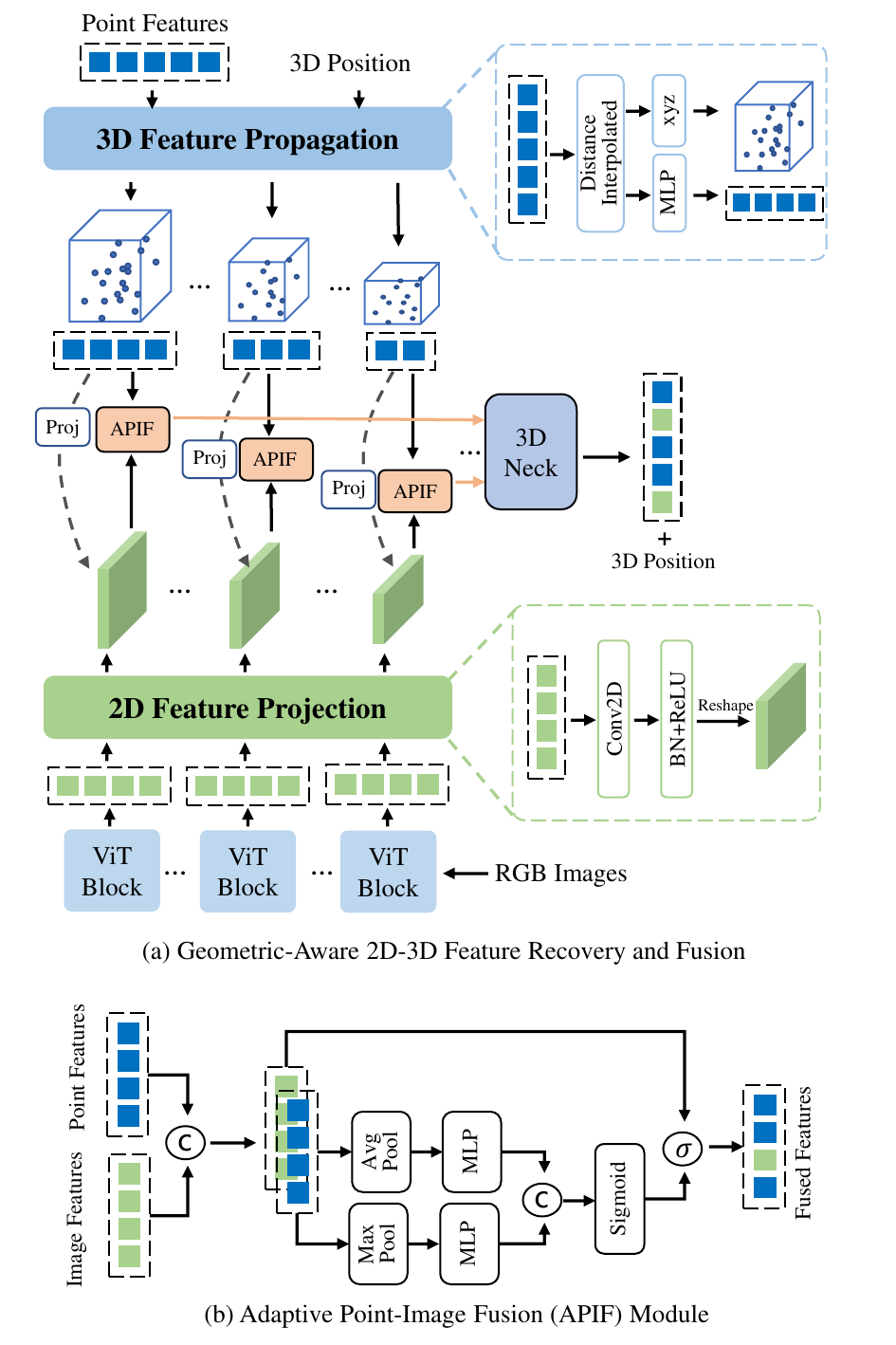}
    \caption{(a) The Geometric-Aware 2D-3D Feature Recovery and Fusion (GARF) is designed to project point-cloud features onto image features and perform fusion. (b) APIF is designed to adaptively fuse point-cloud features with image features.}
    \label{fig:method2}
\end{figure}
After extracting both point cloud and image features using the unified CLIP visual encoder, the challenge lies in effectively fusing these features while preserving the intrinsic 3D spatial information to adapt to the 3D tasks. Traditional 2D feature fusion methods often struggle to maintain the geometric structure of point clouds when integrating them with 2D image features. To address this, as shown in Figure \ref{fig:method2}, we propose the Geometric-Aware 2D-3D Feature Recovery and Fusion (GARF). To preserve the 3D spatial geometric information, we first recover the point cloud sequence features extracted by CLIP into multi-scale 3D sparse features, while the image sequence features are reconstructed into multi-scale 2D feature maps. Subsequently, following the existing 3D-to-2D projection approach\cite{embodiedscan}, the multi-scale 3D sparse features are projected onto the corresponding multi-scale 2D feature maps, obtaining the projected features. These are then fused at multiple scales and passed into the 3D Neck for further fusion and pruning.

\textbf{(1) 3D Feature Propagation.} In this module, we apply 3D feature propagation to recover sparse point-cloud features extracted from the CLIP model. Specifically, given the point-cloud features $f_{p}  \in R^{B\times L \times D_{1}} $ and their corresponding positions, they are passed through the upsampling network, where the input features undergo distance-based interpolation to propagate information between neighboring points. The resulting features are fused with the original point features, and processed through multiple layers of convolution and batch normalization, producing new feature representations. These updated features are then concatenated and fused with additional features to generate the final 3D sparse feature. These processes can be formulated as: 
$$
f_{ps}  = ReLU(BN(Conv(Up(f_{p})))) \eqno{(4)}
$$
where $Up$ represents upsampling. Then, the reconstructed features are concatenated with the original sparse point-cloud features that are input to the CLIP model to construct the Minkowski sparsetensor:
$$
S_{2}  = SparseTensor(Cat(F_{1}, f_{ps}), U_{1}) \eqno{(5)}
$$
Finally, we use 3D Minkowski convolution to generate a multi-scale Minkowski sparsetensor $S^{i} ,i\in (1,2,...,L)$ through $S_{2}$.

\textbf{(2) 2D Feature Projection.} The image features which are extracted by the CLIP visual encoder, are reconstructed into multi-scale 2D feature maps through the 2D Feature Project module. Specifically, we extract the feature $F_{mv} = [f_{mv}^{1} ,f_{mv}^{2},...,f_{mv}^{i}]$, where $ f_{mv}^{i} \in R^{(B* Nums)\times L\times D_{1}}$ from the outputs of multiple layers of the CLIP visual encoder. These features are then processed through 2D convolutional layers with batch normalization and ReLU activation for feature extraction, and finally reshaped into multi-scale 2D image features. These processes can be formulated as below, Where $f_{r}^{i} \in R^{(B* Nums)\times C_{i}\times H_{i} \times W_{i}}$. 
$$
[f_{r}^{1} ,f_{r}^{2},...,f_{r}^{i}] = Reshape(ReLU(BN(Conv(F_{mv})))) \eqno{(6)}
$$
\textbf{(3) Cross-Modal Fusion} After extracting multi-scale 3D sparse point-cloud features and 2D image features separately, the point-cloud features are projected onto the image features using the camera’s intrinsic and extrinsic parameters, forming the projected 3D image sparse features. These features are then fused with the original multi-scale 3D sparse point-cloud features through APIF module to obtain the final image-point cloud fused sparse features. Our APIF module is designed for dynamically fusing multi-scale point-cloud features $F_{s} = [f_{s}^{1} ,f_{s}^{2},...,f_{s}^{i}]$ with the features projected onto the image $F_{proj} = [f_{proj}^{1} ,f_{proj}^{2},...,f_{proj}^{i}]$ . Specifically, Inspired by SENet\cite{senet}, the point-cloud features are first concatenated with the features projected onto the image. After that, the combined features undergo both Max Pooling and Average Pooling operations. These pooled features are then processed through a shared MLP. The outputs from the two pooling operations are concatenated once more and fed into a sigmoid function. Finally, the result from the sigmoid function is multiplied with the initially concatenated features. These processes can be formulated as: 
$$
F_{c} = Cat(F_{s},F_{proj}) \eqno{(7)}
$$
$$
W = Cat(MLP(Maxpool(F_{c})), MLP((Avgpool(F_{c})))) \eqno{(8)}
$$
$$
F_{fuse} = F_{c} \odot sigmoid(W) \eqno{(9)}
$$
Finally, the multi-scale fused features $F_{fuse} = [f_{fuse}^{1},...,f_{fuse}^{i}]$ are fed into the 3D Neck to obtain the final 3D sparse feature $S_{pmv}  = SparseTensor(f_{pmv}, U_{pmv})$, where $f_{pmv}\in R^{N_{s}\times C}, U_{pmv}\in R^{N_{s}\times 3}$.

\subsection{3D Visual-Text Fusion and Decoder}
\begin{figure}[!t]
    \centering
    \includegraphics[width=\linewidth]{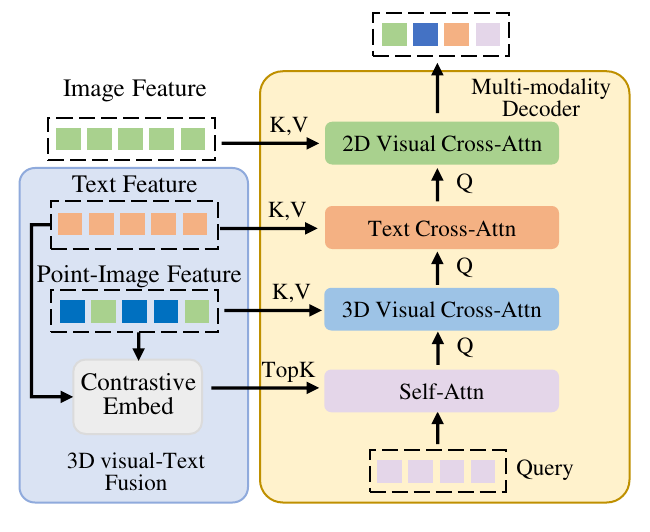}
    \caption{The left part of the figure illustrates the 3D visual-text fusion process, while the right part shows the tri-modal decoder that integrates image, point cloud, and text features for final prediction.}
    \label{fig:method3}
\end{figure}
Effectively extracting key textual cues and integrating them with 3D information is critical for the accuracy of the 3D grounding task. Following EmbodiedScan\cite{embodiedscan}, as shown in the 3D Visual-Text Fusion on the left side of Figure \ref{fig:method3}, a contrastive learning strategy is applied to the fused image-point cloud feature $f_{pmv}\in R^{N_{s}\times C}$ and text feature $f_{t}  \in R^{B\times L \times D_{2} }$ to obtain the final fused feature. Then, the Top-K algorithm is employed to select K features from final fused feature, which are used as queries and fed into the Multi-Modality Decoder.

%%%%%%%%%%%%%%%%%%%%%
\begin{table*}[!t]
\caption{Multi-view 3D object detection benchmark on EmbodiedScan.}
\label{3ddet}
\centering 
\begin{center}
\begin{tabular}{c|cccc|cccccc}
\hline
\multirow{2}{*}{Methods}   &\multicolumn{4}{c|}{Large-Vocabulary}   & \multicolumn{2}{c}{Head} & \multicolumn{2}{c}{Common} & \multicolumn{2}{c}{Tail} \\
    &{$AP_{25}$}    &{$AR_{25}$}   &{$AP_{50}$}   &{$AR_{50}$}    &{$AP_{25}$}    &{$AR_{25}$}   &{$AP_{25}$}    &{$AR_{25}$} &{$AP_{25}$}    &{$AR_{25}$}  \\
\hline
VoteNet \cite{votenet:} & 3.20   & 6.11  & 0.38  & 1.22   & 6.31 & 12.26 & 1.81 & 3.34 & 1.00 & 1.83 \\
\hline
ImVoxelNet \cite{ImVoxelNet} & 6.15   & 20.39  & 2.41  & 6.31   & 10.96 & 34.29 & 4.12 & 15.40 & 2.63 &9.21 \\
\hline
FCAF3D \cite{fcaf3d} & 9.07   & 44.23  & 4.11  & 20.22   & 16.54 & 61.38 & 6.73 & 42.77 & 2.67 & 24.83 \\
+E-decoder\cite{embodiedscan} & 14.80   & 51.18  & 8.77  & 27.46   & 25.98 & 67.12 & 10.85 & 50.08 & 5.72 & \textbf{32.85} \\
+painting \cite{PointPainting} & 15.10   & 51.32  & 8.64  & 26.66   & 26.23 & 67.53 & 11.39 & 50.64 & 5.80 & 32.13 \\
\hline
EmbodiedScan \cite{embodiedscan} & 16.85   & \textbf{51.07}  & 9.77  & \textbf{28.21}   & 28.65 & \textbf{67.51} & 12.83 & \textbf{50.46} & 7.09 & 31.52 \\
\hline
Ours & \textbf{23.37}   & 47.70  & \textbf{13.59}  & 27.02   & \textbf{34.42} & 62.80 & \textbf{19.53} & 46.17 & \textbf{15.62} & \textbf{32.85} \\
\textbf{Improvements} & \textbf{+6.52} &- & \textbf{+3.82} &- & \textbf{+5.77} &- & \textbf{+6.70}& & \textbf{+8.53} & -\\
\hline
\end{tabular}
\end{center}
\end{table*}
%%%%%%%%%%%%%%%%%%%%%
%%%%%%%%%%%%%%%%%%%%%
\begin{table*}[!t]
\caption{Multi-view 3D visual grounding benchmark. “Indep/Dep” refer to “View-Independent/Dependent”. }
\label{grounding}
\centering 
\begin{center}
\begin{tabular}{c|c|cc|cc|cc|cc|cc}
\hline

\multirow{2}{*}{Methods}  & \multirow{2}{*}{Dataset} &\multicolumn{2}{c|}{Overall}   & \multicolumn{2}{c|}{Easy} & \multicolumn{2}{c|}{Hard} & \multicolumn{2}{c|}{Indep} & \multicolumn{2}{c}{Dep}\\
  &  &{$AP_{25}$}    &{$AP_{50}$}   &{$AP_{25}$}    &{$AP_{50}$}   &{$AP_{25}$}    &{$AP_{50}$}   &{$AP_{25}$}    &{$AP_{50}$} &{$AP_{25}$}    &{$AP_{50}$}  \\
\hline
ScanRefer \cite{scanrefer}& EmbodiedScan & 12.85   & -  & 13.78  & -   & 9.12 & - & 13.44 & - & 10.77 & - \\
BUTD-DETR \cite{bottom}& EmbodiedScan & 22.14   & -  & 23.12  & -   & 18.23 & - & 22.47 & - & 20.98 & - \\
L3Det \cite{object2scene}& EmbodiedScan & 23.07   & -  & 24.01  & -   & 18.34 & - & 23.59 & - & 21.22 & - \\
\hline
EmbodiedScan& EmbodiedScan-Mini & 33.59   & 14.40  & 33.87  & 14.58   & 30.49 & 12.41 & 33.61 & 14.65 & 33.55 & 13.92 \\
Ours & EmbodiedScan-Mini & 39.84   & 18.71  & 40.38  & 18.96  & 33.65 & 15.88 & 39.28 & 18.87 & 40.89 & 18.40 \\
\textbf{Improvements} & - & \textbf{+6.25}   & \textbf{+4.31}  & \textbf{+6.51}  & \textbf{+4.38}  & \textbf{+3.16} & \textbf{+3.47} & \textbf{+5.67} & \textbf{+4.22} & \textbf{+7.34} & \textbf{+4.48} \\
\hline
EmbodiedScan& EmbodiedScan-Full & 36.88   & 15.85  & 37.51  & 16.18   & 29.78 & 12.11 & 36.89 & 15.93 & 36.86 & 15.68 \\
Ours & EmbodiedScan-Full & 43.24   & 21.18  &  43.86 &  21.60 & 36.28 & 16.50 & 43.69 & 21.68 & 42.39 & 20.24 \\
\textbf{Improvements} & - & \textbf{+6.36 }   & \textbf{+5.33 }  & \textbf{+6.35 }  & \textbf{+5.42 }  & \textbf{+6.50 } & \textbf{+4.39 } & \textbf{+6.80 } & \textbf{+5.75 } & \textbf{+5.53 } & \textbf{+4.56 } \\
\hline
\end{tabular}
\end{center}
\end{table*}
%%%%%%%%%%%%%%%%%%%%%

Due to the sparsity of point cloud data, small objects are often not effectively captured, which impacts the model’s detection accuracy. To address this limitation, unlike EmbodieScan, we introduce 2D Visual Cross-Attn with query. By fusing image features with point cloud and text features, the detailed information from the image effectively compensates for the deficiencies of point clouds. These processes can be formulated as: 
$$
f_{inter} = CrossAtt(SelfAtt(Q) ,f_{pmv}) \eqno{(10)}
$$
$$
f_{out}=MLP(CrossAtt(CrossAtt(f_{inter},f_{t}),f_{mv}^{1}) ) \eqno{(11)}
$$
The feature $f_{out}$ is then fed into the 3D visual grounding head to output the corresponding object’s 3D bounding box (3D BBOX).
Following EmbodiedScan, we utilize match loss\cite{end} as the loss function for each layer of the decoder. This loss function consists of three components. 
$$
Loss = \alpha L_{Cls}+\beta L_{3DBox}  +\gamma L_{Center}  \eqno{(12)}
$$
Where $L_{Cls}$ represents the contrastive loss applied to queries and textual features for classification, employing Focal Loss\cite{focal} as the method. $L_{3DBox}$ is employed as the L1 loss function for regressing the 9-DOF(Degree of Freedom) 3D bounding box, whereas $L_{Center}$ is utilized for predicting the center point using a cross-entropy loss function. The hyperparameters $\alpha,\beta ,\gamma$ are typically set to 1.0.

\section{Experiments}
In this section, we validate our approach on the tasks of 3D detection and 3D visual grounding, both of which are evaluated based on the EmbodiedScan benchmark.

\subsection{Dataset} 
EmbodiedScan\cite{embodiedscan} is a multi-modal ego-centric 3D perception dataset for embodied AI, comprising 5,185 real-world indoor scans with 890K RGB-D views, 160K oriented 3D bounding boxes, and 970K language prompts. By integrating ScanNet\cite{scannet}, 3RScan\cite{3rscan} and Matterport3D\cite{Matterport3D} dataset with SAM-assisted annotation for small objects and orientation labeling, it supports multi-view 3D perception and 3D visual grounding tasks and there are a total of 284 object categories. For the 3D visual grounding task, EmbodiedScan provides a full dataset consisting of 234,014 3D scene-text pairs, as well as a mini dataset containing 48,120 3D scene-text pairs.

The evaluation protocol adopts average precision metrics computed through 3D Intersection-over-Union(IoU) average precision (AP), employing dual threshold criteria (0.25 and 0.5) to assess performance in both 3D detection and 3d visual grounding tasks. We also use average recall (AR) for reference. For the 3D detection task, we group objects into head, common, and tail types (following EmbodiedScan) and calculate metrics for each group individually. The 3D grounding task utilizes two evaluation dimensions from EmbodiedScan: difficulty categorization, where scenes with over three distracting instances are labeled as challenging, and view sensitivity determination, which marks samples requiring directional text clues (e.g., spatial terms like “front” or “left”) as view-dependent.

\begin{table*}[t]
  \caption{Comparison of model trainable parameters}
  \label{train_para}
    \begin{tabular}{c|cc|cc|cc|cc}
    \hline
    \multirow{2}{*}{Methods}   &\multicolumn{2}{c|}{Encoder}  & \multicolumn{2}{c|}{Decoder} & \multicolumn{2}{c|}{Other}  & \multicolumn{2}{c}{Sum}\\
    &  Num & Size &  Num & Size   &  Num & Size   &  Num & Size      \\
    \hline
    EmbodiedScan & 217.82M & 830.93MB   & 11.60M & 44.23MB  & 0.13M  & 0.51MB  & 229.55M &875.67MB\\
    Ours & 81.83M & 312.16MB   & 14.76M & 56.30MB  & 0.13M  & 0.51MB  & 96.72M & 368.96MB \\ 
    
  \hline
  \end{tabular}
\end{table*}

\subsection{Implementation Details}
The EmbodiedScan dataset consists of RGB-D image sequences, where scene point clouds are generated by projecting multi-view depth images. Specifically, during model training, we select 20 multi-view images with a resolution of 224x224 as input. During testing, we use 50 multi-view images. For each scene, we sample 100,000 points from the original point cloud to serve as the input. We use CLIP Vit-B/16 as the network encoder. The $K$ value is set to 16 in the FPS stage before using CLIP model to extract point-cloud features, the number of point-sample groups is set to 512, and the hidden size is set to 768.
The network is trained using AdamW optimizer\cite{AdamW} with $\beta _{1} =0.9, \beta _{2} =0.999$ and a weight decay of $1e^{-5}$. Our model is trained for 12 epochs each for the 3D detection and 3D visual grounding tasks. The implementation was developed in PyTorch\cite{pytorch}, trained on four NVIDIA L40S GPUs.

\begin{table}[t]
  \caption{Ablation Study on 3D Detection Performance Across Various Datasets}
  \label{diff_data}
  \setlength{\tabcolsep}{3.1pt} 
    \begin{tabular}{c|cc|cc|cc}
    \hline
    \multirow{2}{*}{Methods}   &\multicolumn{2}{c|}{ScanNet}  & \multicolumn{2}{c|}{3RScan} & \multicolumn{2}{c}{Matterport3D} \\
    &{$AP_{25}$}    &{$AP_{50}$}   &{$AP_{25}$}    &{$AP_{50}$}   &{$AP_{25}$}    &{$AP_{50}$}  \\
    \hline
    EmbodiedScan & 22.55 &  12.70  & 18.55 & 9.69  & 10.66  &  6.19 \\
    Ours & 25.18 & 15.23   & 38.11 & 21.58  & 10.74  & 6.24  \\ 
    
    \textbf{Improvements} & \textbf{+2.63}   & \textbf{+2.53}  & \textbf{+19.56}  & \textbf{+11.89}  & \textbf{+0.08} & \textbf{+0.06}\\
  \hline
  \end{tabular}
\end{table}

\begin{table}[t]
  \caption{Ablation Study on 3D visual detection Performance Across Various Datasets}
  \label{diff_data_gd}
  \setlength{\tabcolsep}{3.1pt}
    \begin{tabular}{c|cc|cc|cc}
    \hline
    \multirow{2}{*}{Methods}   &\multicolumn{2}{c|}{ScanNet}  & \multicolumn{2}{c|}{3RScan} & \multicolumn{2}{c}{Matterport3D} \\
    &{$AP_{25}$}    &{$AP_{50}$}   &{$AP_{25}$}    &{$AP_{50}$}   &{$AP_{25}$}    &{$AP_{50}$}  \\
    \hline
    EmbodiedScan & 38.27 &  17.84  & 33.20 & 12.55  &25.78 &  7.41 \\
    Ours & 42.30 & 19.52   & 37.69 & 13.42  & 27.91  & 7.86  \\ 
    \textbf{Improvements} & \textbf{+4.03}   & \textbf{+1.68}  & \textbf{+4.49}  & \textbf{+0.87}  & \textbf{+2.13} & \textbf{+0.45}\\
  \hline
  \end{tabular}
\end{table}

\subsection{Main Results}

\textbf{3D Detection.} Based on the EmbodiedScan benchmark, we selected several different representative models for comparison. VoteNet\cite{votenet:} and FCAF3D\cite{fcaf3d} use depth-projected point clouds as input, while ImVoxelNet\cite{ImVoxelNet} uses only RGB images as input. FCAF3D with painting\cite{PointPainting}, EmbodiedScan\cite{embodiedscan}, and our method all use RGB-D images as input. As shown in Table \ref{3ddet}, it shows the metrics of existing methods, with our method outperforming all others. Compared to EmbodiedScan, our method improves by 6.52\% in the $AP_{25}$ metric and 3.82\% in the $AP_{50}$ metric. Despite this, our method also demonstrates a significant advantage in the less common Tail category.

\noindent \textbf{3D visual grounding.} Table \ref{grounding} presents a detailed comparison between our proposed method and other existing approaches on the EmbodiedScan in 3D visual grounding benchmark. All the proposed methods utilize RGB-D data as input. In previous works\cite{embodiedscan}, they reproduced ScanRefer\cite{scanrefer}, BUTD-DETR\cite{bottom}, and L3Det\cite{object2scene} on the EmbodiedScan benchmark. However, the detailed $AP_{50}$ metrics and the specific dataset usage were not disclosed. Compared to the baseline, our method achieves an improvement of 6.25\% in the combined AP25 metric and 4.31\% in the combined AP50 metric on the EmbodiedScan-mini dataset. Additionally, improvements were observed across all other metrics(e.g. Easy, Hard, Indep) as well.

\noindent \textbf{Model training parameters.} As shown in Table \ref{train_para}, we provide a comparative analysis of the trainable parameters between our model and EmbodiedScan. It is evident that the total number of trainable parameters in our model is only 42.13\% of that in EmbodiedScan. This substantial reduction is largely attributed to our innovative use of the CLIP model to extract features from three modalities (multi-view images, point clouds, and text), which significantly reduces the parameters required in the model encoder.

\begin{table}[t]
  \caption{Ablation Study on GARF for 3D visual grounding}
  \label{GARF}
  \setlength{\tabcolsep}{3.1pt}
    \begin{tabular}{c|cc|cc|cc}
    \hline
    \multirow{2}{*}{Methods}   &\multicolumn{2}{c|}{Overall}  & \multicolumn{2}{c|}{Easy} & \multicolumn{2}{c}{Hard} \\
    &{$AP_{25}$}    &{$AP_{50}$}   &{$AP_{25}$}    &{$AP_{50}$}   &{$AP_{25}$}    &{$AP_{50}$}  \\
    \hline
    Ours w/o GARF & 29.45 & 9.70  & 29.68  & 9.83   & 26.81  & 8.20    \\
    Ours w GARF   & 39.84 & 18.71 & 40.38  & 18.96  & 33.65  & 15.88  \\ 
    \textbf{Improvements} & \textbf{+10.38}   & \textbf{+9.01}  & \textbf{+10.70}  & \textbf{+9.13}  & \textbf{+6.84} & \textbf{+7.68}\\
  \hline
  \end{tabular}
\end{table}

\begin{table}[t]
  \caption{Ablation Study on multi-decoder for 3D visual grounding}
  \label{multi-decoder}
  \setlength{\tabcolsep}{3.1pt}
    \begin{tabular}{c|cc|cc|cc}
    \hline
    \multirow{2}{*}{Methods}   &\multicolumn{2}{c|}{Overall}  & \multicolumn{2}{c|}{Easy} & \multicolumn{2}{c}{Hard} \\
    &{$AP_{25}$}    &{$AP_{50}$}   &{$AP_{25}$}    &{$AP_{50}$}   &{$AP_{25}$}    &{$AP_{50}$}  \\
    \hline
    EmbodiedScan & 33.59   & 14.40  & 33.87  & 14.58   & 30.49 & 12.41 \\
    +Our Decoder & 35.74 & 16.00   & 36.05 & 16.04  & 32.28  & 15.56  \\ 
    \textbf{Improvements} & \textbf{+2.15}   & \textbf{+1.60}  & \textbf{+2.18}  & \textbf{+1.46}  & \textbf{+1.79} & \textbf{+3.15}\\
    \hline
    Ours w/o Decoder & 38.15   & 15.88  & 38.49  & 16.10   & 33.17 & 13.35 \\
    Ours w Decoder & 39.84 & 18.71   & 40.38 & 18.96  & 33.65  & 15.88  \\ 
    \textbf{Improvements} & \textbf{+0.46}   & \textbf{+0.44}  & \textbf{+0.44}  & \textbf{+0.22}  & \textbf{+0.48} & \textbf{+0.85}\\
  \hline
  \end{tabular}
\end{table}

\subsection{Ablation Studies}
Due to the substantial size of the full EmbodiedScan benchmark dataset, we conducted ablation experiments for the 3D visual grounding task exclusively on the mini dataset.

\begin{figure*}[t]
    \centering
    \includegraphics[height=0.25\textheight,width=0.85\textwidth]{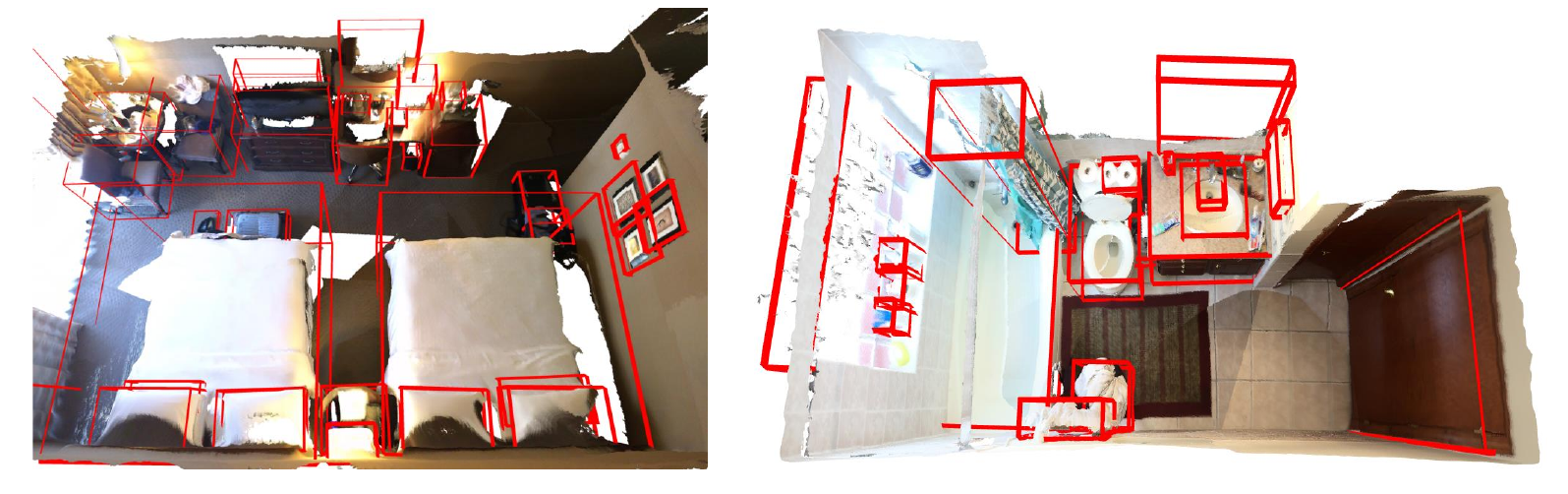}
    \caption{ Qualitative results of 3D detection task}
    \label{fig:vis}
\end{figure*}

\begin{figure*}[t]
    \centering
    \includegraphics[height=0.3\textheight,width=0.9\textwidth]{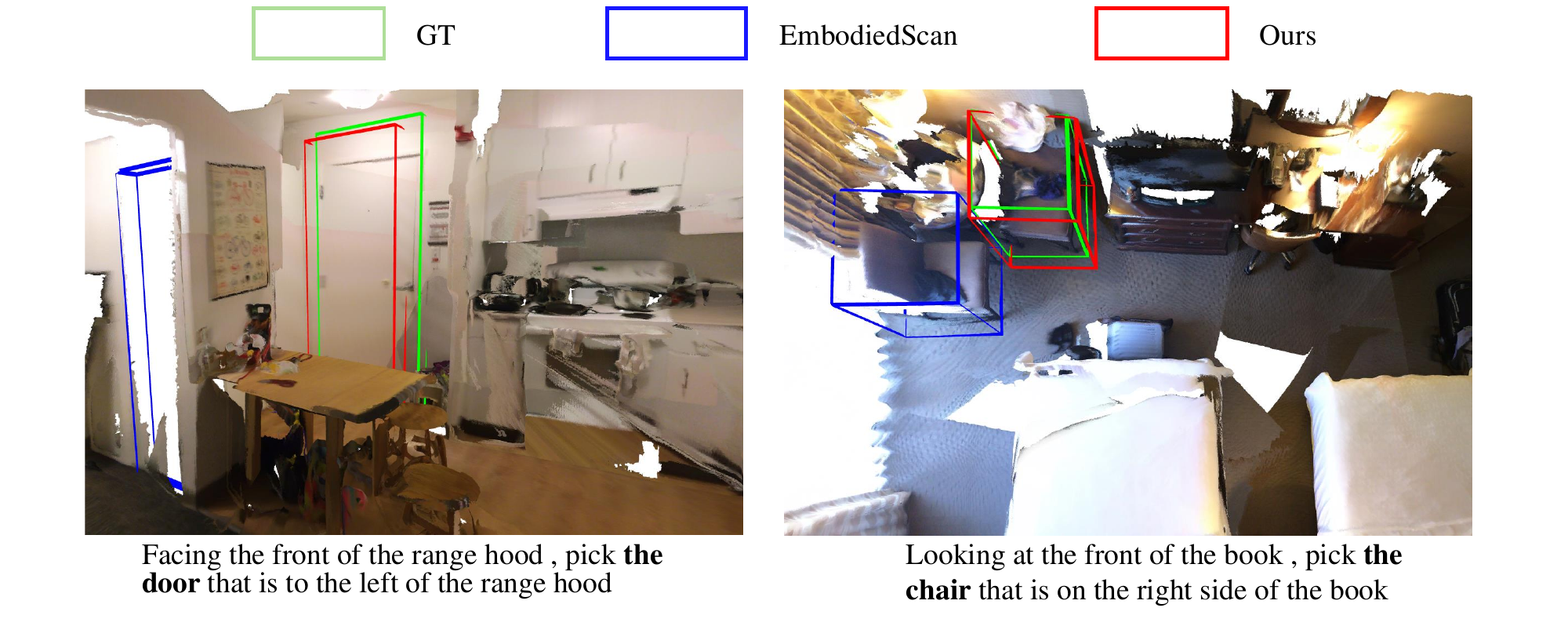}
    \caption{Qualitative results of 3D visual grounding task}
    \label{fig:vis_gd}
\end{figure*}

\noindent \textbf{Different Dataset Ablation.} Since the EmbodiedScan benchmark dataset consists of three sub-datasets (ScanNet, 3RScan, and Matterport3D), we further evaluate the performance of our model on these different datasets, as shown in table \ref{diff_data} and table \ref{diff_data_gd}. Our model achieves improved accuracy across three datasets for both 3D detection and 3D visual grounding tasks. This further validates the advantage of leveraging pre-trained models, which possess strong generalization capabilities.

\noindent \textbf{GARF Module.} Table \ref{GARF} illustrates the effect of our GARF module on the performance of the 3D visual grounding task. It can be observed that integrating GARF into our model results in an accuracy improvement ranging from 6.84\% to 10.70\%. Directly fusing point cloud and image features extracted by CLIP can lead to a loss of corresponding 2D-3D geometric information, resulting in a significant decrease in accuracy. In contrast, our GARF module enhances 3D localization capabilities by reconstructing point cloud-features with 3D information and image feature maps with 2D information. Through multi-scale dynamic fusion, it effectively preserves the geometric information. This approach enhances 3D localization capabilities and improves its accuracy.

\noindent \textbf{Multi-modal Decoder.} To validate the effectiveness of our multi-modal decoder, which integrates 2D features, we conducted ablation experiments on both the EmbodiedScan model and our proposed model. As shown in table \ref{multi-decoder}, our decoder improves accuracy in both models. For EmbodiedScan, the introduction of our decoder results in an accuracy improvement ranging from 1.46\% to 3.15\%. Similarly, in our model, it achieves an accuracy enhancement between 0.22\% and 0.88\%. The introduction of dense 2D features in the multi-modal decoder contributes to a more significant performance enhancement, particularly in Hard scenarios. 

\subsection{Visualization}
To better illustrate the advantages of our method, we present visual comparisons of the network’s predictions. As shown in Fig.\ref{fig:vis}, benefiting from the generalization capability of the CLIP model and GARF module, our network can accurately localize the 3D spatial information of multiple objects within scenes. In Fig.\ref{fig:vis_gd}, the visualization demonstrates that our model achieves more accurate localization compared to EmbodiedScan. This is attributed to our introduction of the CLIP pre-trained model, which enhances its multi-modal understanding capabilities. 

\section{Conclusion}
In this paper, we introduced a novel approach for 3D visual grounding using a unified CLIP-based framework that effectively integrates images, text, and point clouds. Our method simplifies the architecture by eliminating the need for separate 3D networks and employs a Geometric-Aware 2D-3D Feature Recovery and Fusion (GARF) module to enhance cross-modal feature integration. This approach achieves significant accuracy improvements across multiple datasets for both 3D detection and 3D visual grounding tasks. However, challenges remain in optimizing the model’s performance for real-time processing and further reducing computational overhead. Future work will focus on addressing these challenges by enhancing the model’s efficiency and exploring its potential in dynamic, real-world applications.

\section*{ACKNOWLEDGMENT}
This work was supported by the National Natural Science Foundation of China (NSFC) under grants No.62403429, and by Zhejiang Provincial Natural Science Foundation Grant No. LQN25F030008.
\bibliographystyle{ACM-Reference-Format}
\bibliography{Reference}










\end{document}